\documentclass[10pt,letterpaper]{article}

\usepackage{cogsci}
\usepackage{amsmath}

\cogscifinalcopy %

\usepackage{cmap}
\usepackage[T1]{fontenc}
\usepackage[american]{babel}
\usepackage{csquotes}
\usepackage{newtxtext,newtxmath}  %

\usepackage[
  backend=biber,
  style=apa,
  natbib=true,
  eprint=true,
  annotation=false,
]{biblatex}
\usepackage{float} %

\usepackage[hidelinks]{hyperref}

\usepackage{booktabs}
\usepackage{subcaption}
\usepackage{graphicx}

\title{Correlates of Image Memorability in Vision Encoders:\\Activations, Attention Entropy, Patch Uniformity and Autoencoder Losses}

\author{
  {\large\bfseries Ece Takmaz (e.k.takmaz@uu.nl)$^1$ \& Albert Gatt$^2$ \& Jakub Dotla\v{c}il$^1$} \\
  {\normalsize\normalfont
    $^1$Institute for Language Sciences, Department of Languages, Literature and Communication, Utrecht University \\
    $^2$Department of Information and Computing Sciences, Utrecht University 
  }
}

\begin{document}

\maketitle

\begin{abstract}
Images vary in how memorable they are to humans. Inspired by findings from cognitive science and computer vision, we explore correlates of image memorability in pretrained transformer-based vision encoders for the first time. Focusing initially on activations, attention distributions, and the uniformity of image patches, we find that these features correlate with memorability to some extent. Additionally, we explore sparse autoencoder loss over the representations of vision encoders as a proxy for memorability, which yields results outperforming past methods using convolutional neural network representations.
Our results shed light on the relationship between model-internal features and memorability. They show that some features are informative predictors of what makes images memorable to humans; revealing that, in particular, the reconstruction loss from our autoencoders is a strong correlate of image memorability. %

\end{abstract}

\section{Introduction}

\begin{figure}[t]
  \centering
  \begin{subfigure}{0.45\linewidth}
  \hspace{1cm}
  \centering
    \includegraphics[width=0.9\columnwidth]{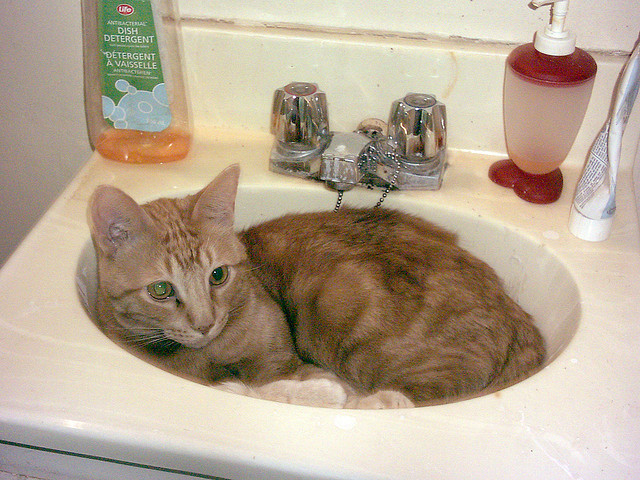}
    \caption{\textbf{Mem:} 0.98 \textbf{Loss:} 294.95}
    \label{fig:short-a}
  \end{subfigure}
  \begin{subfigure}{0.45\linewidth}
  \hspace{0.25cm}\centering
    \includegraphics[width=0.68\columnwidth]{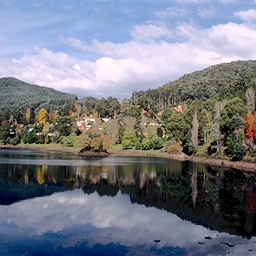}
    \caption{\textbf{Mem:} 0.14 \textbf{Loss:} 219.02}
    \label{fig:short-b}
  \end{subfigure}
  \caption{Images with the highest (a) and lowest (b) memorability scores (Mem) from the MemCat dataset~\citep{goetschalckx2019memcat}, along with the autoencoder reconstruction losses (Loss) incurred by them in our autoencoder whose losses best correlate with image memorability scores.}
  \label{fig:exampleimgs}
  
\end{figure}

Humans can remember very well whether they saw an image, even after viewing 10,000 pictures~\citep{standing10000}, thanks to the
vast storage capacity of visual long-term memory \citep{bradykonkle}. However, some images are more memorable than others; see Figure~\ref{fig:exampleimgs}. \textbf{Image memorability} is a complex phenomenon; yet, it is consistent across individuals, and seems to be an intrinsic property of images~\citep{RUST2020557, BAINBRIDGE20191, Bylinskii2022}, albeit influenced by extrinsic, contextual factors~\citep{extrinsicintrinsic}. 
What makes an image memorable has long been a question that has been posed both in research on visual cognition and computer vision (CV)~\citep{BAINBRIDGE20191, RUST2020557, Bylinskii2022, photomem, whatmakesimg, isola_intrinsic}. %
The former line of research collects data from humans and extracts recognition scores. The latter proposes models to predict memorability and also to edit images to alter their memorability scores. %

Various factors have been identified to be associated with image memorability: \textbf{stronger responses} elicited in the brain for more memorable images~\citep{jaegle}, as well as \textbf{deeper levels of processing} during encoding, which benefits memory retention~\citep{CRAIK1972671}. In CV, scene category is identified as a significant factor in whether an image is memorable--e.g., natural landscape images with \textbf{large uniform regions} are forgotten more than images of people (especially faces) or salient objects~\citep{extrinsicintrinsic, goetschalckx2019memcat, ICCV15_Khosla_lamem, whatmakesimg, photomem}. Conversely, other properties such as interestingness, aesthetics, and low-level features (e.g., color and contrast) are not as correlated with memorability~\citep{ICCV15_Khosla_lamem, photomem, whatmakesimg}. 
The \textbf{distribution of attention} over images is also connected to memorability, with more focused attention observed in more memorable images~\citep{ICCV15_Khosla_lamem, mancas, erdem}. 

Although earlier research used handcrafted features to predict memorability, the introduction of large-scale datasets of images with empirical memorability scores, such as LaMem~\citep{ICCV15_Khosla_lamem} and MemCat~\citep{goetschalckx2019memcat} facilitated the use of deep learning techniques. Influential works mainly use convolutional neural networks (CNNs)~\citep{embracing, compsurvey, squalli, Perera_2019_CVPR_Workshops, ICCV15_Khosla_lamem, resmemnet}, with more recent work also training vision transformers~\citep{hagen2023imagememorabilitypredictionvision, dosovitskiy2021an}. Recently, \citet{Vogelsang2025.09.01.673067} investigated the magnitudes of image representations obtained from layers of CNNs as potential signs of image memorability across a dataset of images that  depict objects \citep[THINGS database;][]{hebart2019things, kramerhebart}.

Closer to our work, \citet{bagheri2025modelingvisualmemorabilityassessment} and \citet{qilin} propose the use of 
\textbf{reconstruction loss from an autoencoder} and \textbf{reconstruction residual from a sparse coding model}, respectively, as a proxy for memorability. Both studies utilize image representations extracted from CNNs and observe  higher reconstruction errors for images that require more complex computations or levels of processing, which are linked to higher memorability~\citep{bagheri2025modelingvisualmemorabilityassessment, qilin}. These approaches carry theoretical and applied relevance. In computational neuroscience, sparse coding has been proposed to explain the sparse activations observed in the primary visual cortex in response to natural images~\citep{olshausen1996emergence, OLSHAUSEN19973311}. In artificial neural network interpretability research, sparse autoencoders have been utilized to represent internal activations as a sparse linear combination of dictionary elements, facilitating the extraction of interpretable features from models~\citep{elhage2022toymodelssuperposition, cunningham, sharkey2023taking}. 

In this paper, for the first time, we investigate the extent to which state-of-the-art \textbf{transformer-based vision encoders} capture \textbf{correlates of image memorability} in a way compatible with the findings in visual cognition, without being trained specifically for this purpose. We first explore the internals of these encoders, inspecting activations, distributions of the attention applied over the image regions and the uniformity of the image regions. Secondly, we train sparse autoencoders built on the representations from vision encoders to investigate the correlation between reconstruction loss and memorability. Our results indicate that model-internal features correlate with memorability to some extent, although the patterns across layers and models are not very consistent. On the other hand, our sparse autoencoders yield losses that strongly correlate with memorability in the MemCat dataset, outperforming previous approaches, as well as generalizing to the LaMem dataset. The features encoded by the autoencoders also allow us to interpret what properties affect the memorability of images the most, supporting previous findings regarding across- and within-scene category variation.\footnote{Code and models available at \url{https://github.com/ecekt/
image_memorability_memvis}}

\section{Methodology}
We explore potential proxies for properties connected to memorability in 2 parts: (1) extracting representations and statistics from state-of-the-art vision encoders and conducting Spearman's correlation analysis and linear regression between model-internal features and memorability; (2) training autoencoders on top of the extracted representations to explore and propose novel setups to investigate the correlation between autoencoder loss and memorability. 

\noindent\textbf{Data.} We primarily use images and memorability scores from the \textbf{MemCat} dataset~\citep{goetschalckx2019memcat}, which consists of 10,000 images each belonging to one of 5 categories: animal, food, vehicle, landscape, and sports. MemCat includes crowdsourced annotations from participants who viewed image sequences and indicated whether a given image had been shown earlier in the sequence. MemCat includes raw data as well as false-recall corrected memorability scores. We use the corrected score, ranging between 0.14 and 0.98, with a mean of 0.693, standard deviation of 0.137 and median of 0.714. Per category means and standard deviations in MemCat are as follows in increasing order of mean: 
\textbf{Landscape}: $\mu= 0.526$, $\sigma= 0.138$; 
\textbf{Vehicle}: $\mu = 0.695$, $\sigma = 0.093$;
\textbf{Sports}: $\mu = 0.713$, $\sigma = 0.098$;
\textbf{Animal}: $\mu = 0.729$, $\sigma = 0.094$; 
\textbf{Food}: $\mu= 0.802$, $\sigma= 0.081$. 

We also utilize \textbf{LaMem}~\citep{ICCV15_Khosla_lamem}, which is a larger dataset consisting of 58,741 images. The dataset provides 5 random split versions consisting of train-validation-test sets with disjoint images and participants. The authors use a random half of the participants for the train and validation scores, and the other half is used to compute the test scores. As a result, an image can have different scores in different versions. To have a single memorability score, we take the average of the scores for an image in all 5 splits. These aggregated scores range between 0.266 and 1.0, have a mean of 0.756, standard deviation of 0.116, and median of 0.769.

\noindent\textbf{Models.} We use the vision encoder from two image-text models, CLIP~\citep{clip} and SigLIP2~\citep{tschannen2025siglip2multilingualvisionlanguage}, as well as a vision-only model DINOv2~\citep{oquab2024dinov}.\footnote{HuggingFace models for CLIP: `openai/clip-vit-base-patch32', SigLIP2: `siglip2-base-patch16-224', DINOv2: `dinov2-base'.}
 All three vision encoders have 12 layers and are based on the Vision Transformer \citep[ViT;][]{dosovitskiy2021an}, which is the architecture of choice for the visual backbone of many state-of-the-art vision-language models, as well as computer vision models. 

One important point to note is that these models were trained with respect to different objectives. CLIP has a contrastive image-text learning scheme, optimizing the alignment between matching images and texts~\citep{clip}. SigLIP2, on the other hand, uses a sigmoid-based learning regime when matching images and texts, in addition to captioning, self-distillation and masked prediction objectives~\citep{tschannen2025siglip2multilingualvisionlanguage}. DINOv2 is not exposed to language, and is trained to predict masked patches in images in a self-supervised manner~\citep{oquab2024dinov}. For CLIP and SigLIP2, supervision with the text encoder was intended to enhance the capabilities in the visual modality. In this way, we explore the effects of having vision encoders trained in coordination with language.  

\citet{darcet2024visionregisters} identify a problem in several vision transformer models including DINOv2, showing the existence of outlier image tokens with very high norms. These tokens, called \textit{registers}, mainly correspond to patches with low information value such as background regions. The authors claim that registers are repurposed to store global image information. Therefore, when we inspect patch uniformity, we also test DINOv3~\citep{simeoni2025dinov3}, which includes extra register tokens to avoid the problem of such artifacts.\footnote{DINOv3-base: `facebook/dinov3-vitb16-pretrain-lvd1689m', DINOv3-large: `facebook/dinov3-vitl16-pretrain-lvd1689m'}

\subsection{Extracting Model-Internal Features}
\label{extractinternal}
Informed by findings from cognitive science and CV, we investigate a set of model-internal features explained below. %

\noindent\textbf{Activations.} In ViTs, each image is divided into a grid of equal-sized patches. The models encode these patches to form image tokens. Prepended to the image tokens is a token called [CLS]. [CLS] captures a `summary' of the image in a vector with 768 dimensions; therefore, we utilize its activations to explore correlates of memorability. We extract the [CLS] representation per layer from the vision encoders. We obtain each [CLS] representation's mean activation, maximum activation, and maximum absolute activation for each of the 12 layers, except the embedding layer, as the initial [CLS] embedding is the same for every image. Note that SigLIP2 does not have a [CLS] token and instead uses a multihead attention pooling head at the final layer. For SigLIP2, we report findings using the first token, though not directly comparable, for compatibility with our layer-wise analysis of the models. One can regard these findings as an analysis of a contextualized image token or a potential \textit{register} token that might capture global image information, as explained earlier~\citep{darcet2024visionregisters}. Our use of [CLS] in what follows includes this first token for SigLIP2. %
Later, we complement this analysis using the pooled output from SigLIP2 to train our autoencoders.

\noindent\textbf{[CLS] delta.} Inspired by the idea that deeper levels of processing lead to more memorability~\citep{CRAIK1972671}, we quantify the magnitude of the change the [CLS] representation goes through, by calculating the cosine distance between [CLS] at adjacent layers. This is a proxy for measuring how levels in image processing relate to memorability.  %

\noindent\textbf{Attention entropy.} As human attention seems to differ based on memorability~\citep{ICCV15_Khosla_lamem, mancas, erdem}, we are 
interested in the distribution of the attention applied by the model to the image patches. The [CLS] token captures global image information by paying attention to patches differently. Therefore, 
we investigate the attention applied by the [CLS] to the image patches. We use the entropy of the attention distribution as a metric that quantifies how spread-out attention is at every layer. 

\noindent\textbf{Patch uniformity.} As less memorable images tend to have no salient, central object~\citep{ICCV15_Khosla_lamem}, we also investigate image patch tokens. Specifically, we compute how varied the image patches of an image are. For each image patch at a given layer, we calculate its average cosine similarity to the rest of the patches for the same image. The mean over all patches of the image gives the image's patch uniformity score. %

\subsection{Autoencoders}
\label{methauto}

Among prior works, \citet{bagheri2025modelingvisualmemorabilityassessment} use  reconstruction losses from a VGG16-based autoencoder fine-tuned on all MemCat images for 1 epoch, using batch size 1, and reaches 0.45 Spearman's correlation coefficient with memorability. \citet{qilin} also explore a sparse coding approach again using VGG16 and report 0.33 Pearson's correlation on the dataset introduced by \citet{photomem}.

We experiment with \textbf{ViTMAE}, a pretrained masked autoencoder based on ViT, which was trained on ImageNet1k~\citep{imagenet}. We use its base and large versions (`facebook/vit-mae-base' and `facebook/vit-mae-large'). We compute autoencoder reconstruction loss for images with 75\% of image patches masked, as this ratio was shown to be optimal~\citep{vitmae}. As MemCat includes 5617 images from ImageNet, we also conduct experiments where we discard this subset and report the correlation for the rest of MemCat ($N=4383$).

Additionally, we train \textbf{our own sparse autoencoders} that take in image representations from CLIP and SigLIP2.\footnote{Using DINOv2 representations as input to our autoencoders yielded only weak correlation ($0.26$), suggesting that language supervision helps capture memorability-related semantics better.} For the CLIP-based autoencoder, we use the [CLS] from the last layer of the vision encoder, while for SigLIP2, we opt for the pooled output as mentioned previously. %

Our autoencoder consists of a linear layer, $W_{enc}$, that projects an image representation, $y$, to a vector of dimensionality 100, ReLU, and a second linear layer, $W_{dec}$, that projects %
back to the dimensions of the image representation:
\begin{equation}
    \hat{y} = \textbf{W}_{dec}(ReLU(\textbf{W}_{enc} y + b_{enc})) + b_{dec}
\end{equation}

We use Mean-Squared Error Loss (MSE) %
as the autoencoder's reconstruction loss and a weighted sparsity penalty (L1 loss) to encourage the model to create sparse latent representations $z$,  constraining the sum of absolute values in the bottleneck of the autoencoder (to aid interpretability, not correlation). In our memorability correlation analyses, we only use the reconstruction loss.

{\begin{equation}
     \mathcal{L} = \underbrace{\sum_{i=1}^{N} (y_i - \hat{y}_i)^2}_{\text{MSE reconstruction loss}} + \underbrace{\lambda \sum_{j=1}^{M} |z_j|}_{\text{sparsity loss}}
\end{equation}}

We use 80\% of MemCat for training and 20\% for validation, after shuffling the whole dataset. Following a hyperparameter search, we train each setup for 5 epochs with batch size 4, learning rate 5e-4, sparsity weight $\lambda =$ 1e-5, and hidden dimensions 100 (best memorability correlations among {50, 100, 200, 500, 768, 1000}). Before training, we apply z-score normalization to image representations per vision encoder. %

\begin{figure*}[]
  \centering
  \begin{subfigure}{0.33\linewidth}
    \includegraphics[width=\columnwidth]{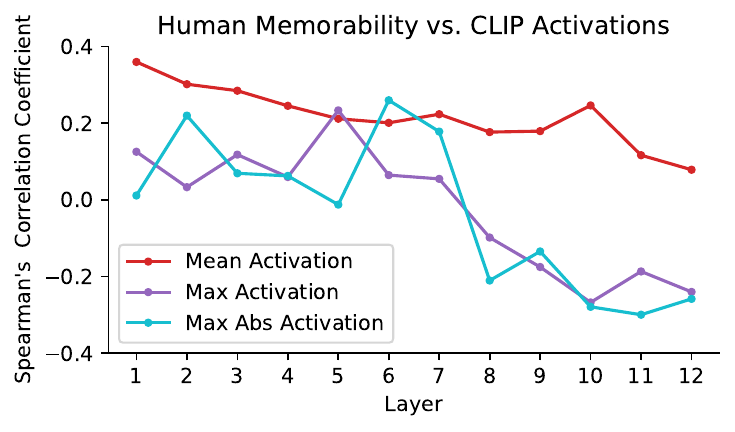}
    \label{fig:short-a}
  \end{subfigure}
  \begin{subfigure}{0.33\linewidth}
    \includegraphics[width=\columnwidth]{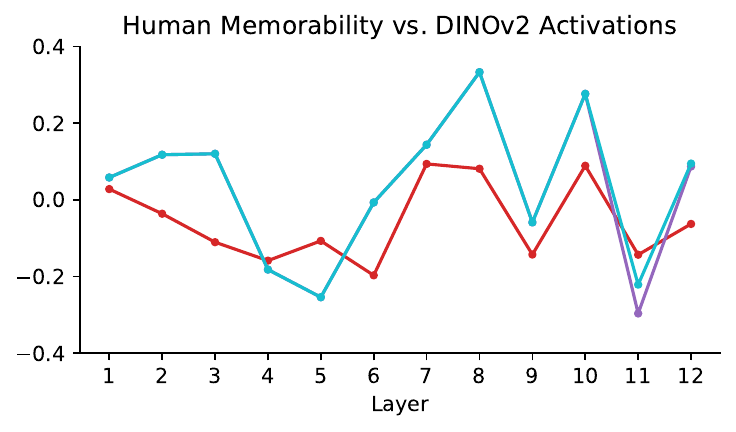}
    \label{fig:short-b}
  \end{subfigure}
  \begin{subfigure}{0.33\linewidth}
    \includegraphics[width=\columnwidth]{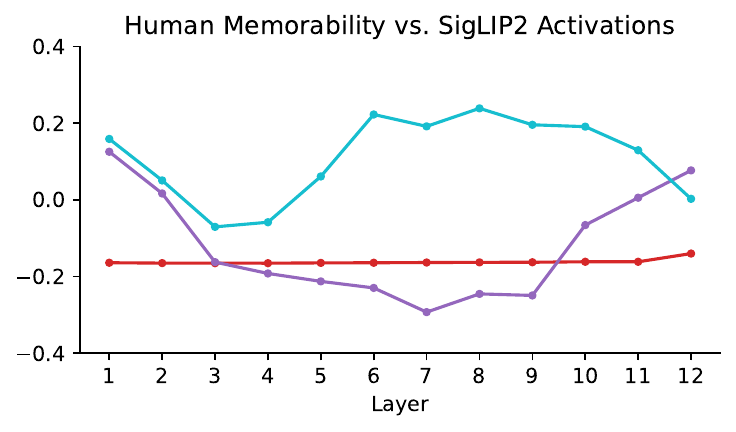}
    \label{fig:short-b}
  \end{subfigure}
  \caption{Correlation coefficients between human memorability and features of [CLS] activations over the layers of vision encoders (except layer 0, because it is the same for all images). \textit{Left:} CLIP, \textit{Middle:} DINOv2, \textit{Right:} SigLIP2.}
  \label{fig:activations}
\end{figure*}

\begin{figure*}[h]
  \centering
  \begin{subfigure}{0.33\linewidth}
    \includegraphics[width=\columnwidth]{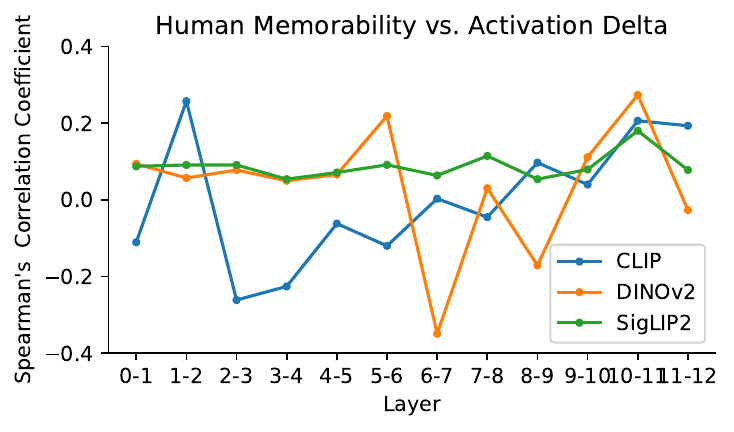}
    \label{fig:short-cls}
  \end{subfigure}
  \begin{subfigure}{0.33\linewidth}
    \includegraphics[width=\columnwidth]{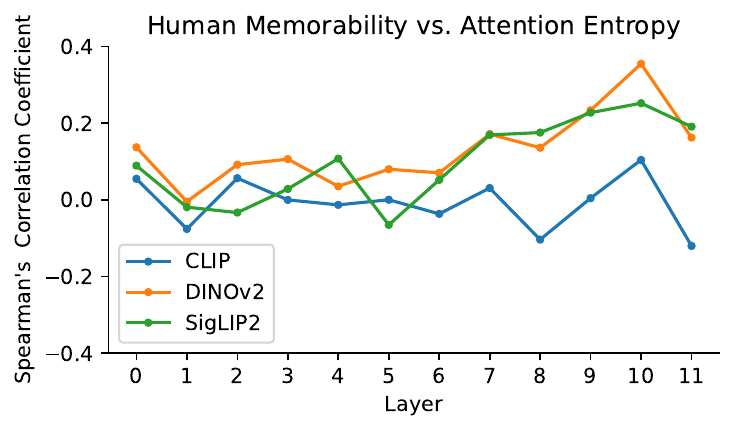}
    \label{fig:short-entropy}
  \end{subfigure}
  \begin{subfigure}{0.33\linewidth}
    \includegraphics[width=\columnwidth]{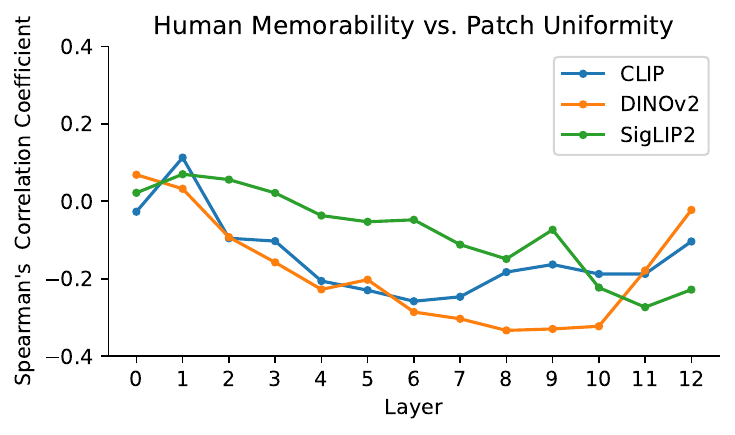}
    \label{fig:short-uniformity}
  \end{subfigure}
  \caption{Correlation between image memorability and \textit{Left:} the cosine distance between adjacent layers' [CLS] activations (delta), \textit{Middle:} entropy of the attention applied by the [CLS] token to the image tokens, and \textit{Right:} patch uniformity.} 
  \label{fig:deltaentruniform}
\end{figure*}

\section{Results}

\subsection{Model-Internal Features}
\label{resactiv}
Figure~\ref{fig:activations} depicts the three encoders' \textbf{activations} in terms of how much they correlate with memorability over the layers. CLIP's mean activation starts from a moderate correlation and displays a trend that goes in the negative direction. DINOv2 Max Abs is a good correlate but fluctuates. SigLIP2's activations are weak correlates, likely because they originated from the first patch. The plot on the left in Figure~\ref{fig:deltaentruniform} depicts that, for CLIP and DINOv2, the deltas show an oscillating correlation to memorability, with the correlation increasing in specific layer intervals and subsequently decreasing, unlike SigLIP2, which has a more stable pattern. %

The middle plot in Figure~\ref{fig:deltaentruniform} illustrates the correlation between \textbf{attention entropy} and memorability. More visibly for DINOv2 and SigLIP2 than for CLIP, we see an increasing trend towards the later layers where the attention distribution reflects information correlating with memorability.

The plot on the right in Figure~\ref{fig:deltaentruniform} reveals the negative correlation in the middle-to-later layers between \textbf{patch uniformity} and memorability. The patch uniformity scores per category are as follows in increasing order: \textbf{Vehicle}: $0.191$, $\sigma = 0.125$;
\textbf{Food}: $0.236$, $\sigma = 0.114$;
\textbf{Sports}: $ 0.257$, $\sigma  = 0.158$;
\textbf{Animal}: $ 0.265$, $\sigma  = 0.151$; 
\textbf{Landscape}: $ 0.354$, $\sigma  = 0.138$. 
In MemCat, landscape images have the lowest memorability as well as higher patch uniformity. This is in line with previous findings about images with no salient and central objects having lower memorability~\citep{ICCV15_Khosla_lamem}. 

\begin{figure}[h]
    \centering
    \includegraphics[width=0.85\linewidth]{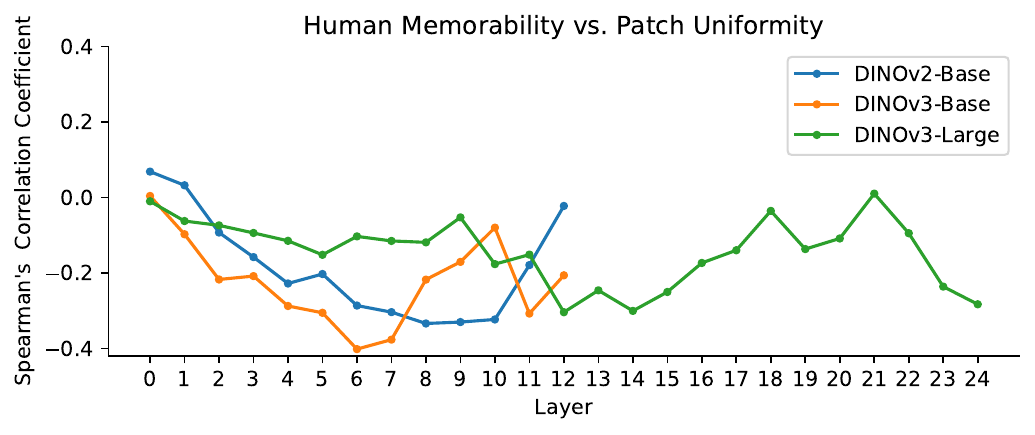}
  \caption{Correlation between patch uniformity and memorability in different versions of the DINO model.}
  \label{dinoregisters}
\end{figure}

Checking different versions of the DINOv3 model, we see that register tokens make patches more meaningful; see Figure \ref{dinoregisters}. DINOv3-Base model has a stronger correlation compared to DINOv2-Base in the first half of the layers, potentially due to the help of the register tokens. DINOv3-Large model yields weaker correlations in the first half, suggesting that large models may be worse at predicting human signals. We continue using DINOv2 for comparability as the other models do not have register tokens.

\begin{table}[h]\footnotesize
\begin{center}
\caption{Linear regression model for DINOv2 layer 10. $^{***}$ indicates significance at $p < .001$. Independent variables were z-score normalized.}
\label{tab:glm}
\vskip 0.12in
\begin{tabular}{lrrr}
\hline
\textbf{}                     & \multicolumn{1}{c}{\textbf{Coef}} & \multicolumn{1}{c}{\textbf{Std Err}} & \multicolumn{1}{c}{\textbf{Z}} \\ \hline
\textbf{Intercept}            & 0.6932                            & 0.001                                & 577.806$^{***}$                         \\
\textbf{Activation max}      & 0.0144                            & 0.001                                & 23.556$^{***}$                          \\
\textbf{Activation mean}     & 0.0264                            & 0.001                                & 20.171$^{***}$                          \\
\textbf{Activation max abs} & 0.0144                            & 0.001                                & 23.556$^{***}$                         \\
\textbf{Patch uniformity}    & -0.0170                           & 0.002                                & -10.574$^{***}$                        \\
\textbf{Attention entropy}  & 0.0457                            & 0.002                                & 27.745$^{***}$                          \\ \hline
\end{tabular}
\end{center}
\end{table} 

We also fit \textbf{linear regression models} to investigate the relationship between model-internal features and memorability as the dependent variable in a single model. To exemplify, in Table~\ref{tab:glm}, we report the outcomes for DINOv2's layer 10 features, as its activations, attention entropy, and patch uniformity show consistently stronger correlations with memorability. The results suggest that activation mean and attention entropy are strong, positive predictors of memorability. On the other hand, as patch uniformity increases, memorability decreases. These findings indicate that, although there does not seem to be a clear pattern among the models and across the layers, there exist model-internal features that correlate with memorability to some extent ($R^2=.238$, for Table~\ref{tab:glm}). 

The variation in these results could be due to the architectures, training sets and training objectives of each model. However, compared to metrics based on activations, the results on two measures reflecting information distribution, i.e., \textbf{attention entropy} and \textbf{patch uniformity} show a clearer pattern across layers, which seem to be valid for all the models.

\subsection{Autoencoder Reconstruction Loss}
\label{resautoenc}
Table \ref{sotaautoencoder} reports the correlation between memorability and reconstruction losses obtained from \textbf{ViTMAE} models. The models yield  very weak correlations, a result that resonates with findings in the linguistic modality:  \textit{larger} pre-trained language models cannot account for human reading signals and diverge from human-like surprisal estimations~\citep{byung}. This could be because these models are also trained on large amounts of data, what is unique or surprising for humans is not so for such models. %
The results are slightly higher, yet still very weak, when we discard the ImageNet data, indicating that these autoencoders cannot be used as off-the-shelf proxies for memorability prediction. 

\begin{table}[h]\small
\begin{center}
\caption{Correlation between image memorability and ViTMAE loss. Very weak correlation, although $p<.001$.}
\label{sotaautoencoder}
    \vskip 0.12in
\begin{tabular}{lcc}
\hline
                     & \textbf{ViTMAE - base} & \textbf{ViTMAE - large} \\ \hline
\textbf{Full MemCat} & 0.073$^{***}$                  & 0.056$^{***}$                   \\ 
\textbf{No ImageNet} & 0.118$^{***}$                  & 0.097$^{***}$                   \\ \hline
\end{tabular}
\end{center}
\end{table}

\begin{figure}[h]
    \centering
    \includegraphics[width=0.75\linewidth]{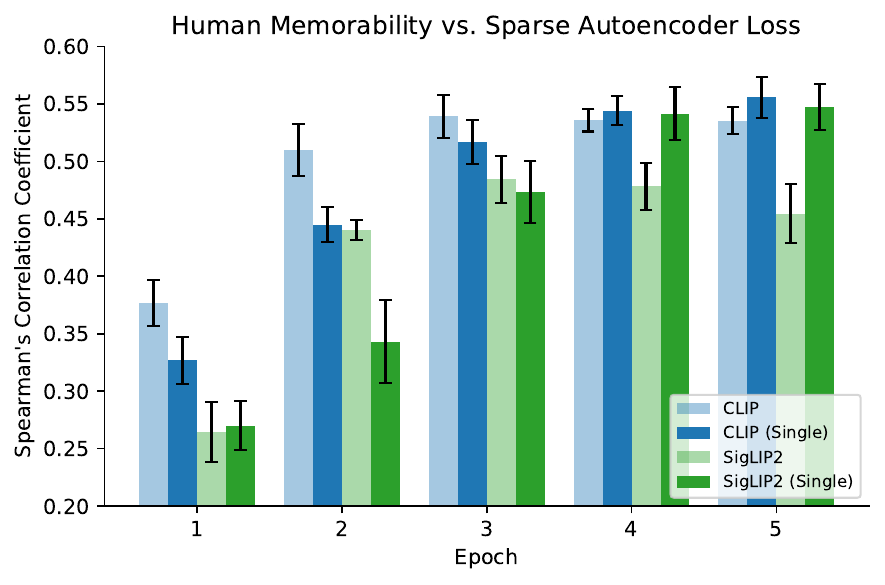}
  \caption{Correlation between reconstruction losses obtained from our sparse autoencoders and memorability on the whole dataset. Error bars indicate standard deviation for 5 random seeds. `Single' is when we train on the whole MemCat with a batch size of 1,  the others follow our initial setting with splits.}
  \label{fig:autoenc}
\end{figure}

On the other hand, the losses from \textbf{our sparse autoencoders} show strong positive correlations with memorability; see Fig.~\ref{fig:autoenc}. CLIP generally correlates better than SigLIP2. At epoch 3, it reaches 0.57 in one run, outperforming previous work using autoencoders~\citep{bagheri2025modelingvisualmemorabilityassessment, qilin}. As done by \citet{bagheri2025modelingvisualmemorabilityassessment}, to emulate human data collection settings, we also experiment with training on the whole MemCat instead of dividing it into splits. We test a single-exposure setup similar to theirs (batch size 1 and learning rate 1e-4). Our best-correlating setup is a SigLIP2-based autoencoder, reaching \textbf{0.58} at epoch 5, after which it drops. %
Using this autoencoder, we obtain reconstruction losses for images. We depict illustrative examples in Figure~\ref{fig:exampleimgs}: the most memorable image in MemCat incurs a loss of 294.954, while the least memorable one incurs 219.022. This is in line with findings showing that higher reconstruction loss is paired with higher memorability. %

\begin{figure}[]
    \centering
    \includegraphics[width=0.8\linewidth]{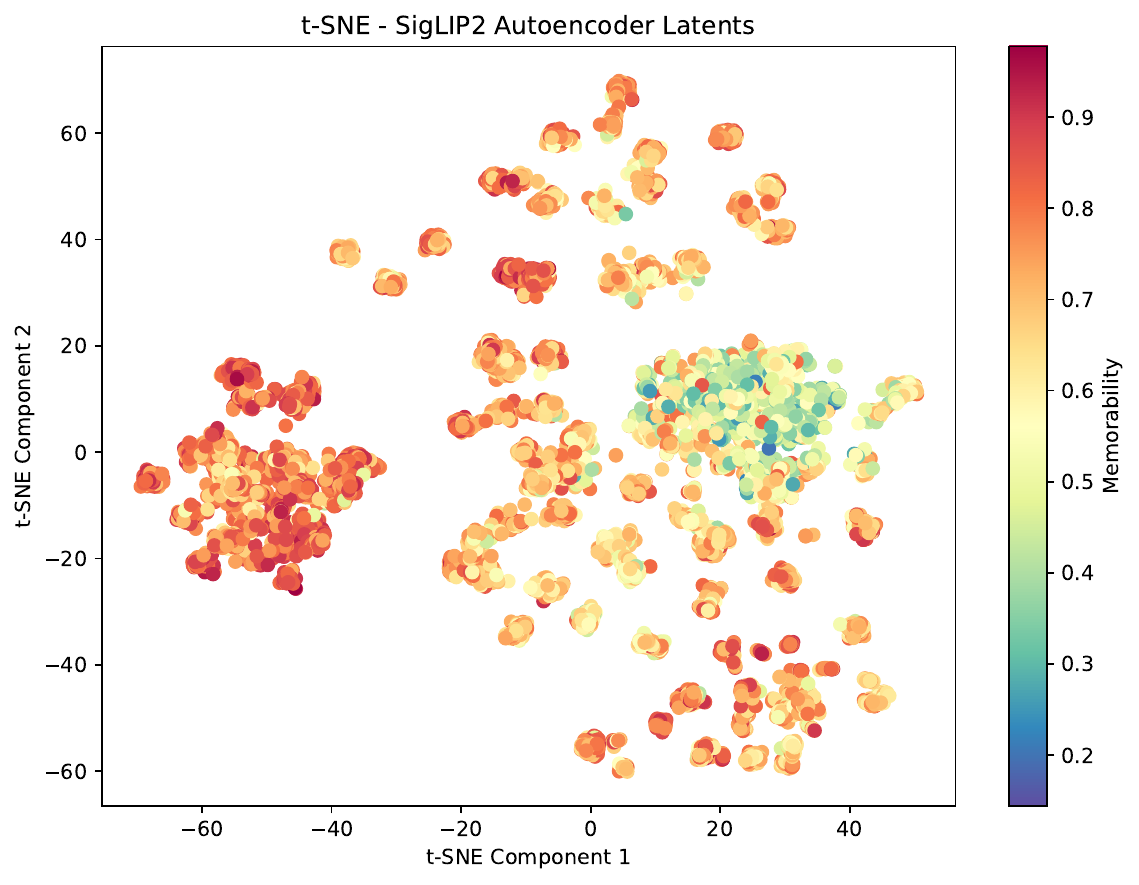}
  \caption{t-SNE visualization of the latent representations from the encoder of the best-correlating SigLIP2-based autoencoder. Red - higher memorability, blue - lower.}
  \label{fig:tsne}
\end{figure}

\begin{figure}[h]
    \centering
    \includegraphics[width=0.7\linewidth]{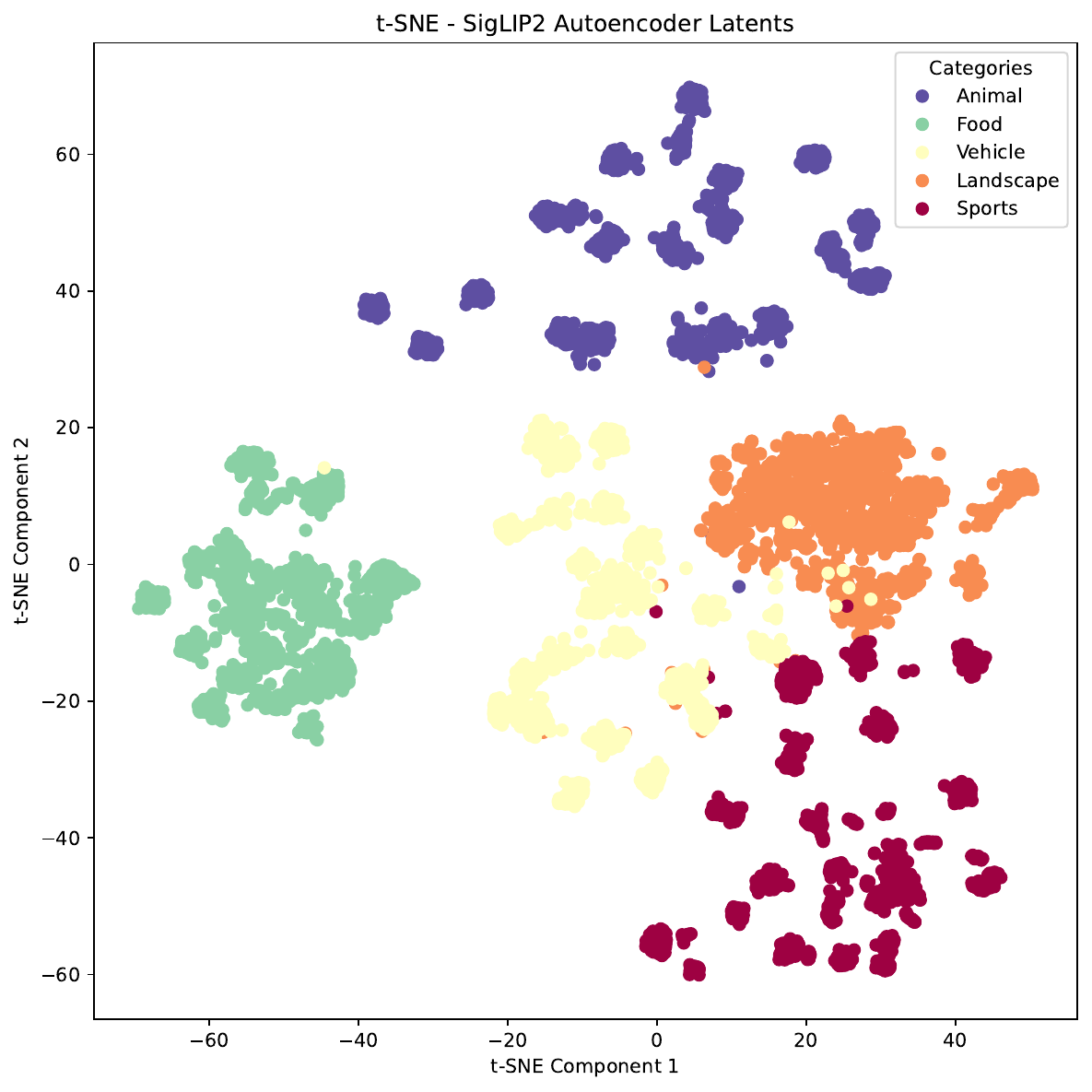}
  \caption{t-SNE visualization of the latent representations from the encoder of the best-correlating SigLIP2-based autoencoder, annotated with image categories.} %
  \label{fig:tsnecat}
\end{figure}

\noindent\textbf{Generalization.} Using the \textit{aggregated} LaMem scores, we test whether our best autoencoder generalizes to images from a different dataset. The best CLIP-based autoencoder yields a correlation of 0.449, and the SigLIP2-based one yields 0.423 correlation, both $p<.001$. We then employ the 5 \textit{test} set splits of LaMem. The best CLIP-based autoencoder reaches 0.414 correlation, and the SigLIP2-based one reaches 0.388 correlation, both $p<.001$. 
The MemNet model proposed together with LaMem~\citep{ICCV15_Khosla_lamem}, which is a CNN finetuned to predict memorability scores, achieves 0.64 correlation. Our results show that our autoencoder losses can generalize and significantly correlate with the memorability scores of novel images from a larger dataset. %

\noindent\textbf{Latent Representation Space. }To have a better picture of the best-correlating autoencoder, we focus on the latent representations extracted from its \textit{encoder}. We find that latents of high-memorability images have larger activations, with mean absolute activations correlating significantly with memorability (0.41 for CLIP, 0.32 for SigLIP2, $p<.001$). In Figure~\ref{fig:tsne}, we illustrate the spatial distribution of the latents, color-coded by their memorability scores. Many clusters consisting of middle and high memorability are spread around the plot. However, the least memorable images seem to be clustered separately. Virtually all of these correspond to landscape images. Figure~\ref{fig:tsnecat} depicts the same plot annotated with image categories, which shows that the images of different categories are well separated in the autoencoder's latent space. 

Within categories, there is some variation in memorability, as previously observed~\citep{extrinsicintrinsic}. As an example, we zoom into the representations of animal images. In the animal cluster, there clearly exist images with low memorability, especially as the image gets closer to the landscape cluster. To check this quantitatively, we compute a mean landscape representation by taking the average of the SigLIP2 representations of images belonging to the landscape category. We find that the closer an animal image's SigLIP2 representations to the mean landscape representation in terms of cosine similarity, the lower its memorability score ($r=-0.452, p<.001$). Additionally, cosine similarity between autoencoder latents of animal images and the mean latent representation of landscape images yields similar results ($r=-0.314, p<.001$).

\begin{figure}[t]
  \centering
  \begin{subfigure}{0.4\linewidth}
  \hspace{1cm}
  \centering
    \includegraphics[width=0.9\columnwidth]{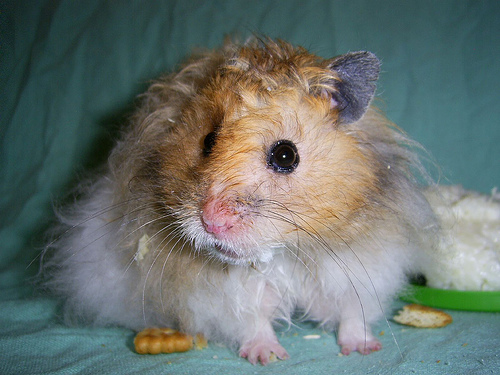}
  \end{subfigure}
  \begin{subfigure}{0.4\linewidth}
  \hspace{0.25cm}\centering
    \includegraphics[width=0.9\columnwidth]{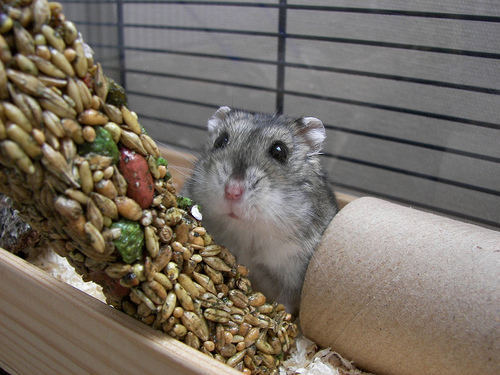}
  \end{subfigure}
  \caption{Images that activate features 97 (left) and 54 (right) the most. These features have the strongest \textbf{positive} correlation  with memorability ($\rho_{97}=0.28, \rho_{54}=0.24$; $p<.001$).}
  \label{fig:featuresactpos}
  
\end{figure}

\begin{figure}[t]
  \centering
  \begin{subfigure}{0.4\linewidth}
  \hspace{1cm}
  \centering
    \includegraphics[width=0.9\columnwidth]{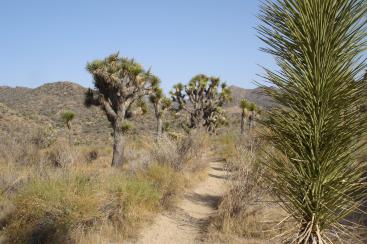}
  \end{subfigure}
  \begin{subfigure}{0.4\linewidth}
  \hspace{0.25cm}\centering
    \includegraphics[width=0.9\columnwidth]{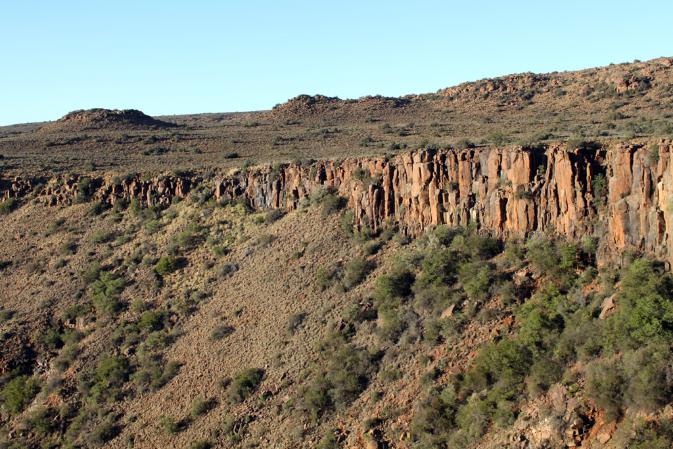}
  \end{subfigure}
  \caption{Images that activate features 63 (left) and 34 (right) the most. The  features have the strongest \textbf{negative} correlation with memorability ($\rho_{63}=-0.49, \rho_{34}=-0.48$; $p<.001$).}
  \label{fig:featuresactneg}
  
\end{figure}

\noindent\textbf{Interpreting Latent Dimensions.} Inspired by studies on the mechanistic interpretability of latent spaces of sparse autoencoders for representations from vision-language models~\citep{papadimitriou2025interpreting, bhalla2024interpreting, stevens2025interpretabletestablevisionfeatures, Olson_2025_ICCV}, we also look deeper into dimensions of the latent representations from the autoencoder. To assess the dimensions, i.e., features, in terms of their relation to memorability, we extract latent representations of all images. Then, we create 100 lists of 10,000 items corresponding to each dimension's activation per image. Afterwards, we compute the correlation between each list and the memorability scores. In this way, we identify the features that significantly correlate with memorability.

Then, we find the images that activate these features the most. We depict the top images for the features with the strongest positive and negative correlations with memorability in Figures~\ref{fig:featuresactpos} and \ref{fig:featuresactneg}, respectively. Features that are activated by images containing beaches, forests, and mountains go together with the lowest memorability scores. On the other hand, features activated by images with objects such as hamsters, limousines, and butternut squash correlate with higher memorability.  We observe the influence of semantic categories on memorability~\citep{kramerhebart, photomem, ICCV15_Khosla_lamem} also in the latents of our autoencoder.

\section{General Discussion}
Image memorability is a multifaceted phenomenon influenced by intrinsic and extrinsic factors, and what makes an image memorable has been studied in cognitive science and computer vision. In this paper, for the first time, we investigated correlates of image memorability in a set of model-internal features obtained from transformer-based vision encoders, which were not trained to predict memorability. We find features that are reasonably good predictors of image memorability. Significantly, reconstruction loss from our sparse autoencoders is a strong predictor. Our findings resonated with those by \citet{qilin} and \citet{bagheri2025modelingvisualmemorabilityassessment}, confirming that difficult-to-reconstruct images are more memorable. Combined with our results on the negative correlation between patch uniformity and memorability, this points to image complexity and distinctiveness as common drivers of both reconstruction difficulty and human memorability.  The outcome regarding the higher activations in the latent representations for memorable images also mirrors the findings about stronger neural responses to memorable images~\citep{jaegle}. We observed across-category and within-category variation in memorability, as reflected in their representations and losses ~\citep{extrinsicintrinsic, photomem, ICCV15_Khosla_lamem, goetschalckx2019memcat}. We also identified and interpreted features that correlate strongly with memorability, which can be further studied to gain insight into fine-grained attributes making an image memorable. Future work can explore a wider range of models, datasets, and other factors linked to memorability such as emotions and context~\citep{Bylinskii2022}. In addition to providing pointers to correlates that would inform models of visual memory, there is potential for transferring our approaches to study processes related to the memorability of other modalities, such as language.

\section{Acknowledgments}

The research reported in this paper was supported by the European Research Council (ERC), grant 101088098 - MEMLANG. Views and opinions expressed are however those of the authors only and do not necessarily reflect those of the European Union or the European Research Council Executive Agency. Neither the European Union nor the granting authority can be held responsible for them. An initial version of this work was presented at the ICCV 2025 Workshop MemVis: The 1st Workshop on Memory and Vision. We thank the reviewers and the audience for their valuable comments.

\printbibliography

\end{document}